\documentclass{article}
\usepackage{ijcai19}

\usepackage{times}
\usepackage{soul}
\usepackage{url}
\usepackage[hidelinks]{hyperref}
\usepackage[utf8]{inputenc}
\usepackage[small]{caption}
\usepackage{graphicx}
\usepackage{amsmath}
\usepackage{booktabs}
\usepackage{algorithm}
\usepackage{algorithmic}
\usepackage{epsfig}
\usepackage{amssymb}
\usepackage{multirow}
\usepackage{bbding}
\usepackage{xspace}
\usepackage{enumerate}
\usepackage{makecell}
\usepackage{caption}
\usepackage{subcaption}

\usepackage{algorithm}
\usepackage{algorithmic}
\usepackage{mathrsfs} % For LaTeX2e
\usepackage{amsmath,bm}
\usepackage{color}
\usepackage{caption}
\usepackage{booktabs}
\usepackage{array}
\usepackage{float}

\def\eg{{\em e.g. }}
\def\ie{{\em i.e. }}
\def\etal{{\em et al. }}

\title{Weakly Supervised Bilinear Attention Network for \\ Fine-Grained Visual Classification}

\author{
 Tao Hu$^1$, Jizheng Xu$^2$, Cong Huang$^2$, Honggang Qi$^1$, Qingming Huang$^1$, Yan Lu$^2$, \\
 $^1$ University of Chinese Academy of Sciences, Beijing, China\\
  $^2$ Microsoft Research Asia, Beijing, China\\
\textit {hutao16@mails.ucas.ac.cn}
}

\begin{document}

\maketitle

\begin{abstract}
  For fine-grained visual classification, objects usually share similar geometric structure but present variant local appearance and different pose. Therefore, localizing and extracting discriminative local features play a crucial role in accurate category prediction. Existing works either pay attention to limited object parts or train isolated networks for locating and classification. In this paper, we propose Weakly Supervised Bilinear Attention Network (WS-BAN) to solve these issues. It jointly generates a set of attention maps (region-of-interest maps) to indicate the locations of object's parts and extracts sequential part features by Bilinear Attention Pooling (BAP). Besides, we propose attention regularization and attention dropout to weakly supervise the generating process of attention maps. WS-BAN can be trained end-to-end and achieves the state-of-the-art performance on multiple fine-grained classification datasets, including CUB-200-2011, Stanford Car and FGVC-Aircraft, which demonstrated its effectiveness.
\end{abstract}

\begin{figure}[t]
	\centering
	\includegraphics[width=0.9\linewidth]{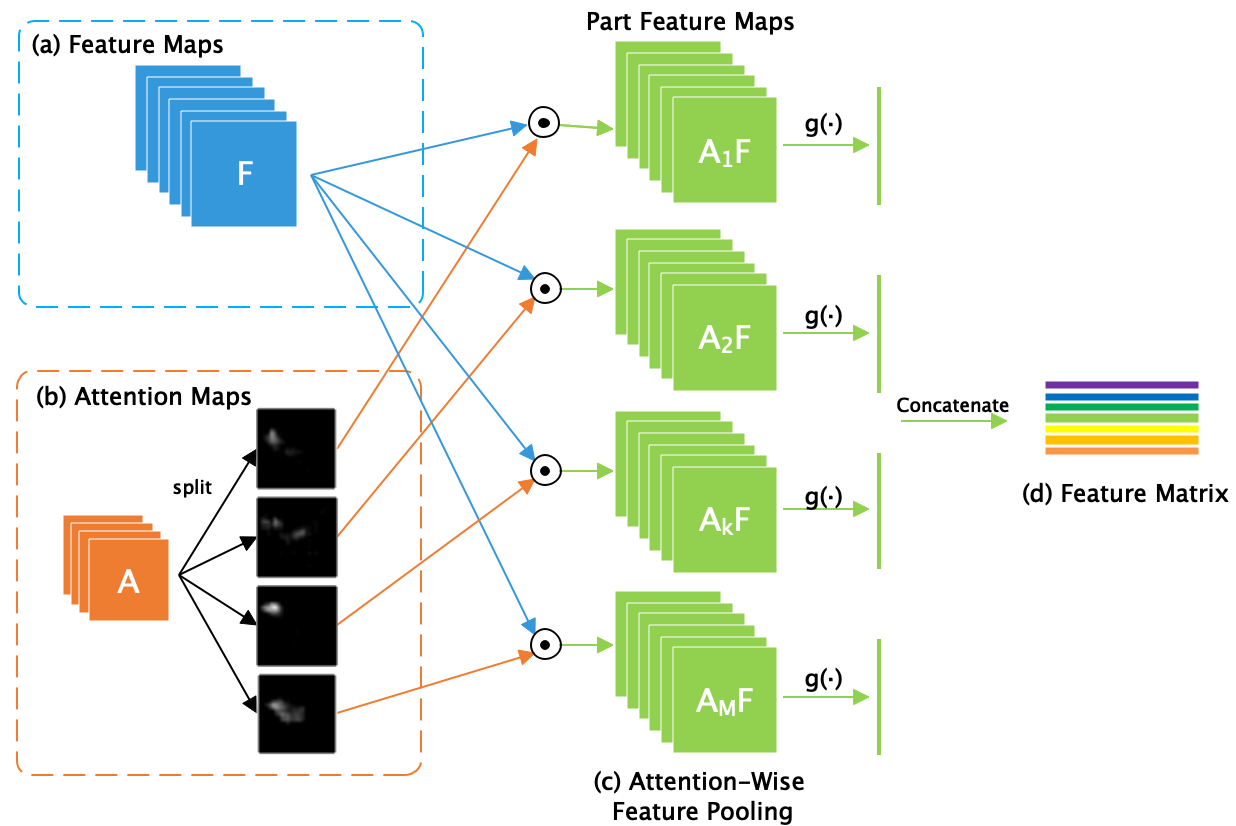}
	\caption{The process of Bilinear Attention Pooling. The network backbone first generates feature maps (a) and attention maps (b). Each attention map represents one of object's visual parts; (c) Part feature maps are generated by element-wise multiplying each attention map with feature maps. Then, part features are extracted by convolution and pooling operation $g(\cdot)$; (d)The final feature matrix is obtained by concatenating all these part features.}
	\label{fig:attention_pooling}
\end{figure}

\section{Introduction}
\label{sec:introduction}
Fine-Grained Visual Classification (FGVC) aims to classify subordinate level categories under some basic-level category, such as species of the bird, model of the car, type of the aircraft and identity of the human face.
Although great success has been achieved for basic-level classification in the last few years, such as ImageNet~\cite{imagenet}, by using Convolutional Neural Network (CNN),  FGVC is still challenging because of three main reasons: (1) High intra-class variances. Objects that belong to the same category usually present significantly different poses and viewpoints; (2) Low inter-class variances. Objects that belong to different categories may be very similar apart from some minor differences, \eg color style of a bird's head can usually determine its category; (3) Expensive human annotation results in limited training data. Labeling fine-grained categories usually require specialized knowledge and a large amount of annotation time.

To distinguish fine-grained categories that have very similar features, the key process is to focus on the feature representation of the object's parts. According to whether the method requires additional part location annotation, current state-of-the-art methods can be divided into two groups, \ie image-level annotation~\cite{stn,bcnn,racnn,MA-CNN,PSM:AAAI2017} and location-level annotation~\cite{MA-CNN}. In the training process, the former only requires image category labels while the latter demands additional location information, such as the location of the bounding boxes or key-points. Location annotation brings about more expensive human labeling cost, which makes it harder to be implemented. As a consequence, researchers pay more attention to classify fine-grained categories only by image-level annotation. In fine-grained classification task, image-level based methods first predict the location of object's parts by taking full advantage of the image category, and then extract corresponding local features. Our work also follows this stream.

However, existing image-level based methods for fine-grained classification usually suffer from two issues: (1) They usually predict locations of a small number (1 to 8) of object's parts, such as bird's head, beak and wings, which limits the accuracy. For the reason that once some of these parts are invisible or occluded, part features will be incorrectly extracted and the classification result is very likely to be wrong; (2) Many methods trained only by softmax cross entropy loss tend to pay major attention to the most discriminative parts(\eg bird's head) and ignore the less discriminative parts(\eg bird's belly), which results in inaccurate object localization and feature representation. The problem is also illustrated in \cite{racnn}.

In this work, inspired by bilinear pooling~\cite{bcnn}, we propose Weakly Supervised Bilinear Attention Network (WS-BAN) to deal with the above issues, which jointly learns the location of a large number of discriminative object's parts and their feature representation to improve the accuracy. As mentioned in~\cite{Zhang2016PickingDF}, convolutional layer usually matches with a potential type of geometric distribution or visual pattern. Our object's parts are represented by attention maps instead of bounding boxes, and the part features are extracted by Bilinear Attention Pooling (BAP), which makes the model feasible to be trained end-to-end.

We expect that $i_{th}$ attention map to represent $i_{th}$ object's part. Nevertheless, if there is no constraint, attention maps tend to be sparse and random, as illustrated in Fig.~\ref{fig:visualization_lan}. In our method, we avoid the problem by introducing weakly supervising attention learning. For attented part features that belong to the same category, we proposed attention regularization to make sure that each part feature will get close to a its part center, as illustrated in Fig.~\ref{fig:center_loss}. To prevent attention maps from mainly focusing on the most significant object's parts and ignoring other less discriminative parts, we propose attention dropout, which randomly drops out some attention maps during training so as to provide the possible situations that any object's part is invisible and enhance the activated value of the less discriminative parts.

Since our attention maps represent the visual parts' distribution of objects, we can use them to locate objects in images. A simple strategy is averaged the attention maps along the channel direction to obtain the object mask and set a threshold to calculate the bounding box. The benefit of object localization is that we can enlarge the object and inference its category to further refine the classification probability.

Our contributions can be summarized as follows:
\begin{enumerate}
	\item We propose Bilinear Attention Pooling to jointly pay attention to a set of object's parts and pool discriminative local features. Based on BAP, we construct a network called Weakly Supervised Bilinear Attention Network to solve the challenging fine-grained visual classification problem.
	\item We weakly supervise attention maps learning process by attention regularization and attention dropout. Attention regularization supervises $k_{th}$ attention map to represent the same $k_{th}$ object's part, and attention dropout encourages the model to pay attention to more discriminative object's parts.
	\item  The number of attention maps can be easily increased and we can utilize them to accurately locate object and further improve the classification accuracy.
  We conduct comprehensive experiments on polular fine-grained visual classification datasets and achieve the state-of-the-art performance. Experiments about attention regularization, attention dropout and number of attention maps also demonstrate the effectiveness of WS-BAN.
\end{enumerate}

The rest of the paper is organized as follows. Firstly, we review related work of WS-BAN in Section~\ref{sec:related_work}. Secondly, we describe the proposed WS-BAN in detail in Section~\ref{sec:approach}, including bilinear attention pooling and weakly supervised attention learning. Thirdly, comprehensive experiments are performed to demonstrate the effectiveness of WS-BAN in Section~\ref{sec:experiments}. Finally, conclusion is drawn in Section~\ref{sec:conclusion}.

\begin{figure*}[t]
	\centering
	\includegraphics[width=0.7\linewidth]{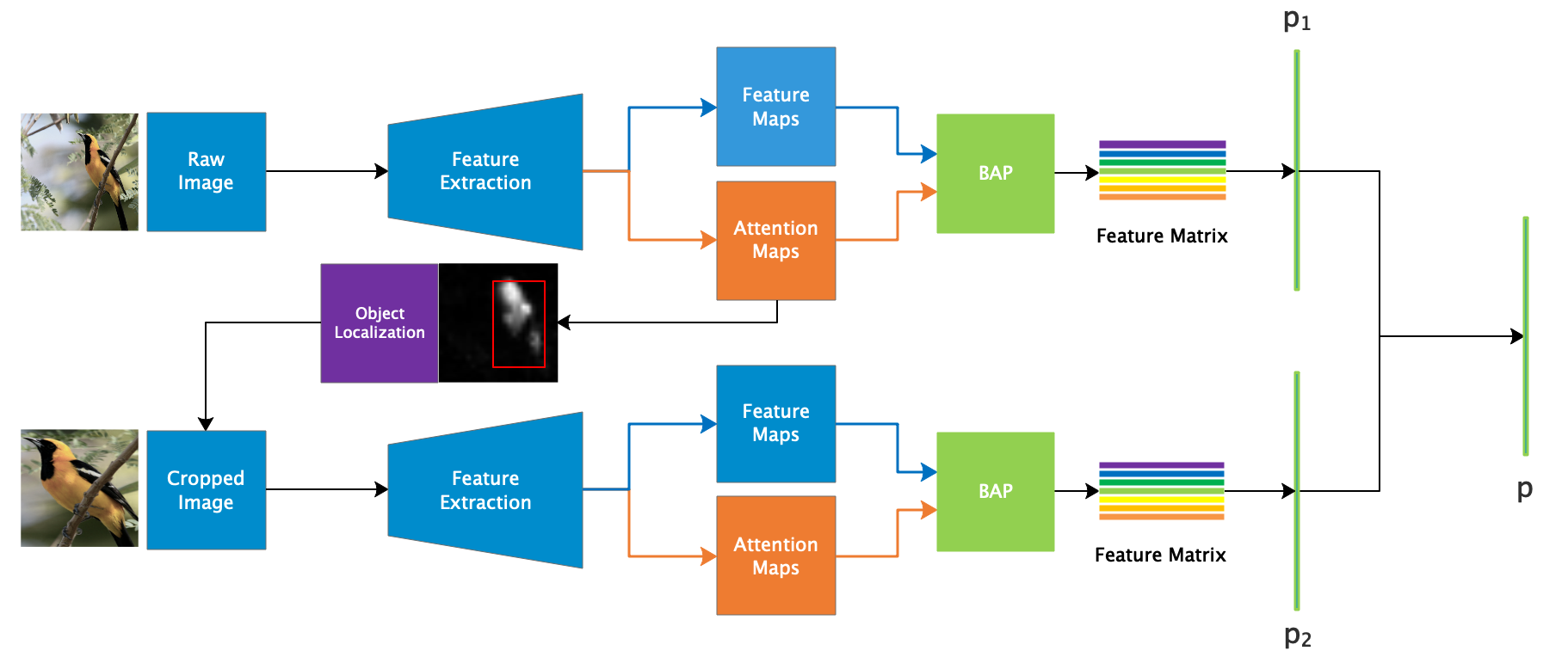}
	\caption{\small Overview of the proposed Weakly Supervised Bilinear Attention Network (WS-BAN). Part feature matrix is generated by Bilinear Attention Pooling (BAP) as illustrated in Fig.~\ref{fig:attention_pooling}. Attention regularization and attention dropout are proposed for the attention map learning process, which supervises each attention map to focus on one of the object's parts. For fine-grained classification task, part feature matrix is normalized and vectorized before outputting the category probability $p_1$. In particular, the mean attention maps can be utilized to segment the foreground from the background and then predict the location of the object. The region of object can enlarge to further refine the probability $p_2$ in the next stage. These two stages share all parameters and the final result is the mean value of their category probabilities.}
	\label{fig:overview_framework}
\end{figure*}

\section{Related Work}
\label{sec:related_work}
In this section, we review the related work about fine-grained visual classification, object localization, bilinear pooling and visual attention.
\paragraph{Fine-grained Visual Classification}
A variety of methods have been proposed to distinguish different fine-grained categories.

Convolutional Neural Networks, such as VGGNet~\cite{vgg}, ResNet~\cite{resnet} and Inception~\cite{inception}, were proposed to solve the large scale image classification problem. While according to the experiments results in Table~\ref{tab:cub}, these basic models can only achieve the moderate performance, because without special design, it is difficult for them to focus on the subtle differences of object's parts.

To focus on the local features and reduce the human labeling cost, many methods that only require image-level annotation has been proposed in recent years. Spatial Transformer Network (ST-CNN)~\cite{stn} aims to achieve more accurate classification performance by first learning a proper geometric transformation and align the image before classifying. The method can also locate several object's parts at the same time.
Similar with ST-CNN, Fu \etal proposed Recurrent Attention CNN (RA-CNN)~\cite{racnn} to recurrently predict the location of one attention area and extract the corresponding feature, while the method only focuses on one local part at a time, and they combine three scale features to predict the final category.
To generate multiple attention locations in the same time, Zheng \etal proposed Multi-Attention CNN (MA-CNN)~\cite{MA-CNN}, which simultaneously locates several body parts. And they proposed channel grouping loss to generate multiple parts by clustering. However, the number of object's parts is still limited (2 or 4), which might constrain their accuracy. In Section~\ref{sec:experiments} Table~\ref{tab:num_parts}, we demonstrate that classification accuracy can be improved by introducing more object's parts.

\paragraph{Weakly Supervised Object Localization}
Weakly supervised learning is an umbrella term that covers a variety of studies that attempt to construct predictive models by learning with weak supervision \cite{review:wsl}, which mainly consists of incomplete, inexact and inaccurate supervision. Localizing object or its parts only by image-level annotation belongs to the inexact supervision.

Accurately Locating the object or its parts only by image-level supervision is very challenging. Early works~\cite{box_predict,Zhang2016TopDownNA} usually generate class-specific localization maps by Global Average Pooling. The activation area can reflect the location of an object. However, training by softmax cross entropy can only output bounding box just covers part of the object. To locate the whole object. Zhang \etal proposed Adversarial Complementary Learning (ACoL) ~\cite{acol} approach to discover entire objects by training two adversary complementary classifiers, which can locate different object's parts and discover the complementary regions that belong to the same object.

\paragraph{Bilinear Pooling}
Bilinear pooling (BP)~\cite{bcnn} aggregates the location-wise outer-product of two features by global averaging, as represented in Equ.~\ref{equ:bp}
\begin{equation}
B = \dfrac{1}{L} \sum_{i=1}^{L} f_A^i {f_B^i} ^T
\label{equ:bp}
\end{equation}
where $L$ is the number of feature locations and $f_A^i, f_B^i$ are two feature representations in $i_{th}$ location.

Our Bilinear Attention Pooling is inspired by BP, which also extracts feature representation by combining features from two streams. However, we treat one stream as feature maps, and supervise another stream as attention maps to indicate parts' distribution of object. The final feature is concatenated by these attention features.

\paragraph{Visual Attention}
Visual attention module~\cite{residual_attention_network,learning_attention} has been proposed to pay more attention to target region by assigning different weights to spatial locations. Previous works usually insert single visual attention module between adjacent layers, which might limits the performance. Our proposed method also belongs to visual attention while the number of our visual attention module are easily to be extended, which is proved to be beneficial to improve the accuracy, as shown in Table~\ref{tab:num_parts}.

\section{Approach}
\label{sec:approach}
In this section, we describe the proposed WS-BAN in detail, which consists of BAP, weakly supervised attention learning and post-processing for classification and object localization.
The overall network structure is illustrated in Figure~\ref{fig:overview_framework}.

\subsection{Bilinear Attention Pooling}
% Similar with bilinear pooling which aggregates features from two streams, we combine two convolutional layers to obtain feature representation. However, we explicitly treat one convolutional layer as feature maps and supervise the other convolutional layer to represent the spatial distribution of object's part.

We generate feature maps $F\in R^{H\times W\times N}$ and attention maps $A\in R^{H\times W\times M}$ by one or several convolutional operations from the convolutional neural network backbone, note that both $F$ and $A$ have the same map size $H \times W$. Attention maps are then split into M maps $ A=\{a_1, a_2, ..., a_M\} $. We expect that $a_k$ to reflect the region of $k_{th}$ object's part.

After that, we element-wise multiply feature maps $F$ by each attention map $a_k$ in order to generate $M$ part feature maps $F_k$, as shown in Equ~\ref{equ:attention_pooling},

\begin{equation}
\begin{aligned}
F_k = a_k \odot F \quad(k = 1,2,...,M)
\end{aligned}
\label{equ:attention_pooling}
\end{equation}
where $\odot$ indicates element-wise multiplication for two tensors.

Similar with Landmarks-Attention Network~\cite{sir}, we further extract discriminative local feature by additional local feature extraction function $g(\cdot)$, such as Global Average Pooling (GAP), Global Maximum Pooling (GMP) or convolutional operation, in order to obtain $k_{th}$ part feature representation $f_k\in{R^{1 \times N}}$,
\begin{equation}
f_k = g(F_k)
\end{equation}

The final part feature matrix $P \in R^{M \times N} $ is concatenated by these local features. In summary, let $\Gamma(A, F)$ indicates the Bilinear Attention Pooling between attention maps $A$ and feature maps $F$. $P$ can be represented by Equ~\ref{equ:label_matrix},
\begin{equation}
P= \Gamma(A, F)
= \begin{pmatrix} g (a_1 \odot F) \\ g(a_2 \odot F) \\ ... \\ g(a_M \odot F)  \end{pmatrix}
= \begin{pmatrix}f_1 \\ f_2 \\ ... \\ f_M \end{pmatrix}
\label{equ:label_matrix}
\end{equation}

\begin{algorithm}[h]
	\begin{algorithmic}[1]
		\REQUIRE Feature Maps $F\in R^{H \times W \times N}$
		\REQUIRE Attention Maps $A\in R^{H \times W \times M}$
		\STATE $ P \gets  \O$ \\
		\FOR {$a_k$ in $A$}
		\STATE Part feature maps pooling: $F_k \gets a_k \odot F$\\
		\STATE Local feature extracting: $f_k \gets g(F_k)$ \\
		\STATE Append part feature: $P \gets P \bigcup f_k$
		\ENDFOR
		\RETURN $P$
	\end{algorithmic}
	\caption{Bilinear Attention Pooling}
	\label{alg:sampling}
\end{algorithm}

% In particular, it can be easily proved that when $g(\cdot)$ is global average pooling, $\Gamma(A, F)$ can be simplified as bilinear pooling
% \begin{equation}
% P = \Gamma(A, F) = A^TF = \dfrac{1}{L} \sum_{k=1}{L} a_i f_f^T
% \end{equation}

\subsection{Weakly Supervised Attention Learning}
Learning attention maps is one of the most important process in achieving high accuracy, since only image-level annotations are available. We propose attention regularization and attention dropout to weakly supervise the process.
\subsubsection{Attention Regularization}
\begin{figure}[h]\
	\centering
	\includegraphics[width=0.7\linewidth]{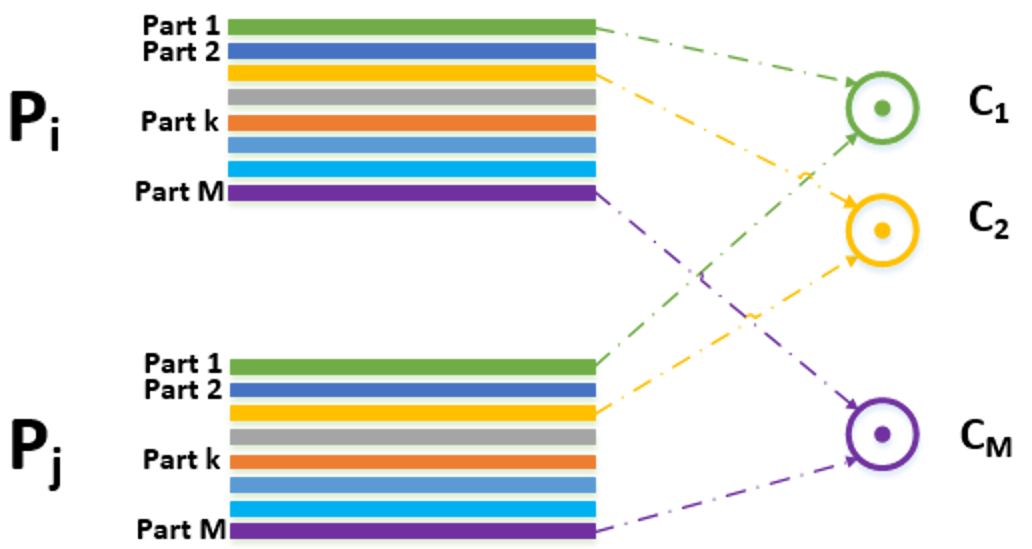}
	\caption{The extract part feature matrix $P_i, P_j$ belong to the same fine-grained category. They consist of $M$ part feature vectors. We weakly supervise its part features $f_k$ that belong to the same part to get close to a part feature center $c_k$.}
	\label{fig:center_loss}
\end{figure}

We expect that $k_{th}$ attention map represents $k_{th}$ object's part. However, in practice, training only by softmax cross entropy loss cannot gurantee that. Inspired by center loss~\cite{center_loss} proposed to solve face recognition problem, we propose attention regularization. Attention maps learning process is weakly supervised by penalizing the variance of features that belong to the same object's part, which means that part feature will get close to the a feature center as illustrated in Fig~\ref{fig:center_loss}. The regularization loss can be represented by $L_A$ in Equ~\ref{equ:attention_regularization_loss},
\begin{equation}
L_{A} = \sum_{k=1}^M \| f_k - c_k \|_2^2
\label{equ:attention_regularization_loss}
\end{equation}
where $c_k$ is $k_{th}$ feature center for each category. $c_k$ is initialized from zero and updated by moving average,
\begin{equation}
c_k \gets c_k + \beta (f_k - c_k)
\end{equation}
where $\beta$ controls the update rate of part center $c_k~\in R^{1\times N}$.

\subsubsection{Attention Dropout}
Attention maps tend to be significantly activated in the most discriminative parts, which results in over-fitting problem. In this paper, inspired by dropout~\cite{dropout} which randomly drops neurons in order to provide an effective way to combine exponentially different neural networks, we propose attention dropout to disperse attention. Attention maps are randomly dropped with a fixed probability $(1 - p)$ ($p$ is the keep probability) during training. When the most discriminative parts are dropped and ignored, the network is forced to enhance the activated value of less discriminative parts, which increase the robustness of classification and accuracy of object localization. Attention dropout can be represented by Equ~\ref{equ:attention_dropout},
\begin{equation}
a_k \gets (m \cdot a_k) / p
\label{equ:attention_dropout}
\end{equation}
where $m \sim Bernoulli(p)$.

\begin{figure}[h]
\centering
\includegraphics[width=0.5\linewidth]{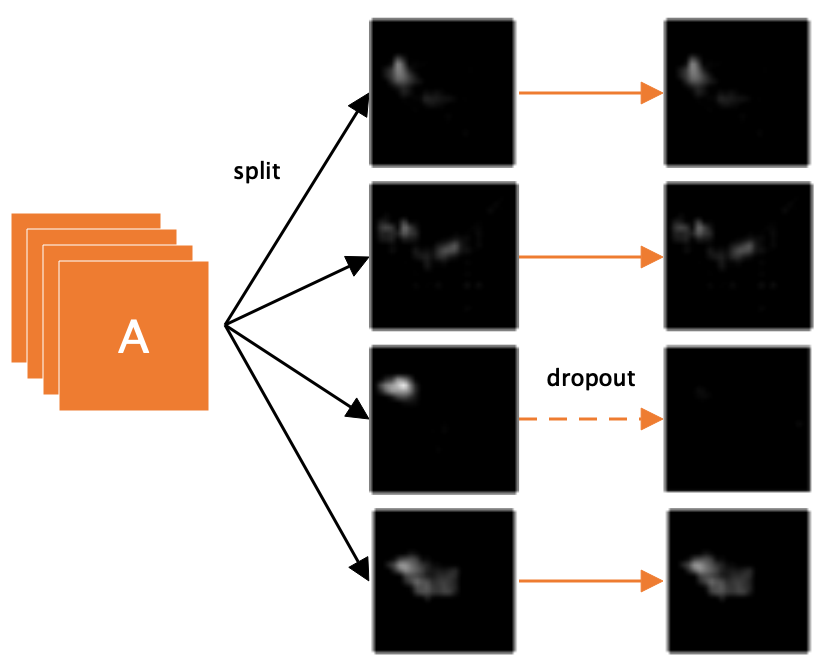}
\caption{Attention Dropout during training. Each attention map is randomly dropped in the training process, which provides the all the possible situations that any object's part is invisible and prevent the model from only focusing on the most significant object's parts. }
\label{fig:attention_dropout}
\end{figure}

\subsection{Object Classification and Localization}
Based on the discriminative part feature matrix $P$ and attention maps $A$, we can solve the fine-grained visual classification and object localization problems.
\subsubsection{Fine-grained Visual Classification}
For classification task, we first normalize $P$ by~\cite{feature_normalization}, i.e. sign-sqrt normalization ($y=sign(x) \cdot \sqrt{\left| x \right |}$) followed by $l_2$ normalization ($z=y/ \| y \|_2 $). Then additional feature vectorizing, fully convolution and softmax operation are employed to predict the probability of each category.

\subsubsection{Object Localization}
Attention maps represent different parts of object. To predict the location of the whole object, we first average attention maps along channel direction to obtain object mask $S\in R^{H\times W}$,
\begin{equation}
\label{equ:mean_map}
S = \dfrac{1}{M} \sum_{i=1}^M a_k
\end{equation}
Then, we apply the same strategy detailed in~\cite{box_predict} to predict the object bounding box based on the localization (attention) maps and then segment the foreground from the background by a fixed threshold $\theta$. Finally, we find a bounding boxes which can cover the foreground pixels.

\subsubsection{Two-stage Refinement}
To further improve the accuracy of classification. After calculating the location of object. We crop and resize the region of object as the input for the second stage, then it is inferenced to refine the prediction, as illustrated in Figure~\ref{fig:overview_framework}. These two stages share the same parameters and the final result is mean value of their category probabilities.

The final loss function consists of softmax cross entropy loss and attention regularization loss, as represented in Equ~\ref{equ:final_loss}.
\begin{equation}
L = L_1 + L_2 + \lambda L_A
\label{equ:final_loss}
\end{equation}
where $L_1$ and $L_2$ are softmax cross entropy loss in two stages respectively, $L_A$ is attention regularization loss (Equ~\ref{equ:attention_regularization_loss}) and $\lambda$ controls the ratios of them.

\section{Experiments}
\label{sec:experiments}
In this section, we show comprehensive experiments to verify the effectiveness of WS-BAN. Firstly, we show the contributions of our proposed components. Then, we compare our method with the state-of-the-art methods on three publicly available fine-grained visual classification datasets. Following this, we explore the effect of attention maps's number. Finally, we visualize attention maps intuitively.
\subsection{Datasets}
\paragraph{Datasets} We compare our method with the state-of-the-art methods on three FGVC datasets, including CUB-200-2011~\cite{CUB_200_2011}, Stanford Cars~\cite{Stanford_car}, and FGVC-Aircraft~\cite{fgvc_aircraft}. Specific information of each dataset is shown in Table~\ref{tab:dataset}.

\begin{table}[h]
	\begin{center}
		\scriptsize
		\begin{tabular}{c|c|c|c|c}
			\hline
			Dataset & Object & \#Category & \#Training & \#Testing\\
			\hline
			CUB-200-2011 & Bird & 200 & 5994 & 5794  \\
			\hline
			FGVC-Aircraft & Aircraft & 100 & 6667 & 3333\\
			\hline
			Stanford Cars & Car & 196 & 8144 & 8041 \\
			\hline
		\end{tabular}
	\end{center}
	\caption {Introduction to three common fine-grained visual classification datasets.}
	\label{tab:dataset}
\end{table}

\subsection{Implement Details}
In the following experiments, we adopt Inception-V3~\cite{inception} as the backbone network and choose \textit{Mix6e} as feature maps. Attention maps are obtained by $1 \times 1$ convolutional operation from feature maps. The weight of attention regularization $\lambda$ is set to 1.0 and attention dropout factor $p$ is set to $0.8$, i.e. we randomly drop $20\%$ attention maps. The update rate of each part center $\beta$ is 0.05.

During training, raw images are first resized into $512 \times 512$ then randomly cropped into size $448 \times 448$. We train the models using Stochastic Gradient Descent (SGD) with momentum of $0.9$, epoch number of $80$, weight decay of $0.00001$, and a mini-batch size of $16$ on one P100 GPU. The initial learning rate is set to $0.001$, with exponential decay of $0.9$ after every $2$ epochs.

\subsection{Performance Contribution}
As described above, Our WS-BAN mainly consists of bilinear attention pooling, attention regularization, attention dropout, and two-stage refinement. We perform an experiment in CUB-200-2011 dataset and explore how much contribution each component can make, as shown in Table~\ref{tab:components}. The accuracy of object localization is represented by mIoU (intersection-over-union) of the bounding box with the ground truth object bounding boxes. We can see that they all improve the accuracy of classification and localization a lot.
\begin{table}[h]
    \begin{center}
        \scriptsize
        \begin{tabular}{c|c|c|c|c|c}
            \hline
            \makecell{Attention \\ Pooling} & \makecell{Attention \\ Regularization} & \makecell{Attention \\ Dropout} & \makecell{Refinement} & Accuracy(\%) & mIOU(\%)\\
            \hline
             & & & & 83.7 & 47.8\\
             \checkmark & & & & 84.7 & 53.5\\
             \checkmark & \checkmark & & & 87.0 & 55.2\\
             \checkmark & \checkmark & \checkmark & & 87.5 & 58.2\\
             \checkmark & \checkmark & \checkmark & \checkmark & 88.8 & 58.2\\
            \hline
        \end{tabular}
    \end{center}
    \caption {Contribution of proposed components in CUB-200-2011 bird dataset. The backbone network is Inception-V3.}
    \label{tab:components}
\end{table}

\subsection{Comparison with State-of-the-art Methods}
\paragraph{Classification Results}We compare our method with state-of-the-art baselines on above mentioned fine-grained classification datasets. The results are respectively shown in Table~\ref{tab:cub}, Table~\ref{tab:fgvc} and Table~\ref{tab:car}. Our WS-BAN achieves the state-of-art performance on all these fine-grained datasets. In particular, we significantly improve the accuracy compared with the backbone Inception-V3.

% We perform experiments on three fine-grained classification datasets with the same hyper-parameters and all of them achieve the state-of-the-art performance, which demonstrated that WS-BAN is a adaptive solution for different datasets.

\begin{table}[h]
	\begin{center}
		\scriptsize
		\begin{tabular}{c|c}
			\hline
			Methods  & Accuracy(\%)\\
			\hline
			% Part-RCNN~\cite{Part-rcnn} & \checkmark & 76.4\\
			% DeepLAC~\cite{deeplac} & \checkmark & 82.3\\
			% PA-CNN~\cite{pa-cnn} & \checkmark &  82.8\\
			% MG-CNN~\cite{mg-cnn} & \checkmark & 83.0\\
			% FCAN~\cite{fcan} & \checkmark &  84.3\\
			% B-CNN~\cite{bcnn} & \checkmark &  85.1\\
			% MASK-CNN~\cite{MA-CNN} & \checkmark &  85.4\\
			VGG-19~\cite{vgg} & 77.8 \\
			ResNet-101~\cite{resnet} & 83.5\\
			Inception-V3~\cite{inception} & 83.7\\
      \hline
			% TLAN~\cite{tlan} &  77.9\\
			% MG-CNN~\cite{mg-cnn} &  81.7\\
			% FCAN~\cite{fcan} &  82.0\\
      PA-CNN~\cite{krause_wo_annotation} &  82.8\\
			B-CNN~\cite{bcnn} & 84.1\\
			ST-CNN~\cite{stn} & 84.1\\
			% PDFR~\cite{PDFR} &  84.5\\
			RA-CNN~\cite{racnn} &  85.4\\
			MA-CNN~\cite{MA-CNN} & 86.5\\
      MAMC~\cite{mamc} & 86.5 \\
      PC~\cite{pairwise_confusion} & 86.9 \\
      MPN-COV~\cite{mpn-cov} & 88.7\\
			\hline
			%Inception-V3~\cite{inception} & 83.7 \\
			% \textbf{LAN} & \textbf{85.5} \\
			\textbf{WS-BAN}  & \textbf{88.8}\\
			\hline
		\end{tabular}
	\end{center}
	\caption {Comparison with state-of-the-art methods on CUB-200-2011 testing dataset. }
	\label{tab:cub}
\end{table}

\begin{table}[h]
	\begin{center}
		\scriptsize
		\begin{tabular}{c|c}
			\hline
			Methods  & Accuracy(\%)\\
			\hline
			% MG-CNN~\cite{mg-cnn} & \checkmark & 86.6\\
			% MDTP~\cite{MDTP} & \checkmark &  88.4\\
      VGG-19~\cite{vgg} & 80.5 \\
			ResNet-101~\cite{resnet} & 87.2\\
			Inception-V3~\cite{inception} & 87.4 \\
			\hline
			% FV-CNN~\cite{fv-cnn} &  77.9\\
			B-CNN~\cite{bcnn} & 84.1\\
			RA-CNN~\cite{racnn}  &  88.4  \\
      PC~\cite{pairwise_confusion} & 89.2 \\
			MA-CNN~\cite{MA-CNN} & 89.9  \\
      MPN-COV~\cite{mpn-cov} & 91.4\\
      % MAMC~\cite{mamc} & 93.0 \\
			\hline
			% \textbf{LAN} & \textbf{89.5} \\
			\textbf{WS-BAN}  & \textbf{92.3}\\
			\hline
		\end{tabular}
	\end{center}
	\caption {Comparison with state-of-the-art methods on FGVC-Aircraft testing dataset. }
	\label{tab:fgvc}
\end{table}

\begin{table}[h]
	\begin{center}
		\scriptsize
		\begin{tabular}{c|c}
			\hline
			Methods & Accuracy(\%)\\
			\hline
			% R-CNN~\cite{r-cnn} & \checkmark & 88.4\\
			% FCAN~\cite{fcan} & \checkmark &  91.3\\
			% MDTP~\cite{MDTP} & \checkmark &  92.5\\
			% PA-CNN~\cite{pa-cnn} & \checkmark &  92.8\\
      VGG-19~\cite{vgg} & 85.7\\
			ResNet-101~\cite{resnet} & 91.2\\
			Inception-V3~\cite{inception} & 90.8 \\
			\hline
			RA-CNN~\cite{racnn} & 92.5\\
			MA-CNN~\cite{MA-CNN} & 92.8  \\
      PC~\cite{pairwise_confusion} & 92.9 \\
      MAMC~\cite{mamc} & 93.0 \\
      MPN-COV~\cite{mpn-cov} & 93.3\\
      \hline
			\textbf{WS-BAN} & \textbf{93.6}\\
			\hline
		\end{tabular}
	\end{center}
	\caption {Comparison with state-of-the-art methods on Stanford Cars testing dataset. }
	\label{tab:car}
\end{table}

\paragraph{Object Localization Results}
 The recent method ACoL~\cite{acol} provided the performances of image-based object localization on CUB-200-2011 bird dataset. To compare with them, we evaluate our method on the same dataset with the same metric, \ie calculating the localization error (failure percentage) of the images whose bounding boxes have less 50\% IoU with the ground truth.

 The experimental results are shown in Table~\ref{tab:localiztion_error}. With the contribution of BAP and attention learning, our method surpasses the state-of-the-art methods by a large margin, which shows that our method can pay attention to object's parts correctly.

%localiztion_error
\begin{table}[h]
	\begin{center}
		\scriptsize
		\begin{tabular}{c|c}
			\hline
			Method & Localization Error(\%)\\
			\hline
			GoogLeNet-GAP  &  59.0\\
			VGGnet-ACoL~\cite{acol} &  54.1\\
			\textbf{WS-BAN} & \textbf{28.2}\\
			\hline
		\end{tabular}
	\end{center}
	\caption {Object localization error on CUB-200-2011 testing set.}
	\label{tab:localiztion_error}
\end{table}

\subsection{Effect of Number of Attention Maps}
More object's parts usually contribute to better performance. Similar conclusion was made in MA-CNN~\cite{MA-CNN} and ST-CNN~\cite{stn}. We performed experiments on how the number of attention maps affects the final accuracy. Table~\ref{tab:num_parts} shows the relation between the number of object's parts, classification accuracy, and mIOU on CUB-200-2011 datasets. We can see, with the increasing of number of object's parts, the object classification accuracy and object localization mIOU also rise. When the number of object's parts reaches to around 64, the performance gradually becomes stable, and the final accuracy reaches to 88.8\% and mIOU reaches to 58.2\%. Our end-to-end feature pooling model makes it easily to set arbitrary number of object's parts, we can achieve a more accurate result by increasing the number of attention maps.

\begin{table}[h]
	\begin{center}
		\scriptsize
		\begin{tabular}{c|c|c}
			\hline
			\# Parts & Accuracy(\%) & mIoU(\%)\\
			\hline
			4 & 85.9 & 53.4  \\
      8 & 87.2 & 56.1 \\
      16 & 87.6 & 57.3\\
			32 & 88.3 & 57.8\\
			64 & 88.7 & 58.1\\
      128 & 88.8 & 58.2 \\
      \hline
		\end{tabular}
	\end{center}
	\caption {The effect of number of object's parts evaluated on CUB-200-2011 testing dataset. }
	\label{tab:num_parts}
\end{table}

\subsection{Visualization of Attention Maps}
\begin{figure}[t]
	\begin{subfigure}{.45\textwidth}
		\centering
		\includegraphics[width=\linewidth]{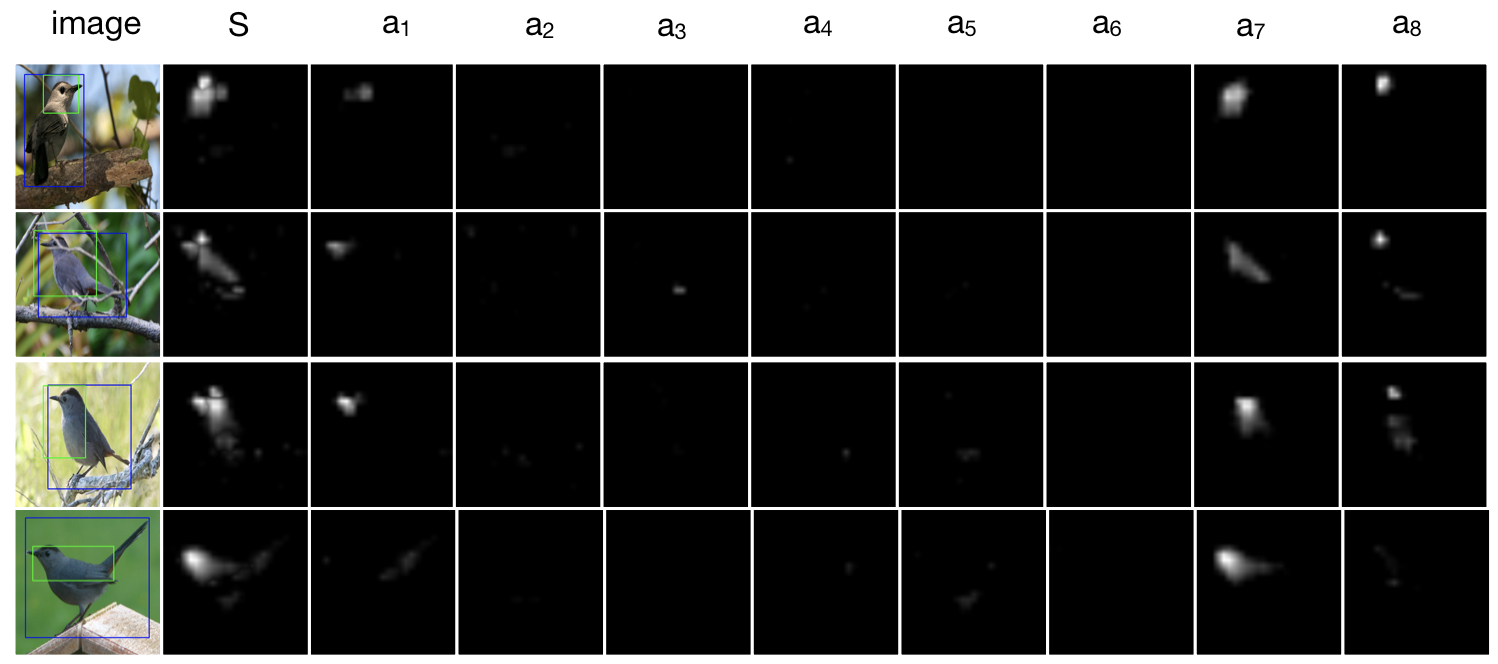}
		\caption{Attention maps generated only by Bilinear Attention Pooling. Without attention regularization and attention dropout, attention maps are sparse and most of then focus on limited object's parts. }
		\label{fig:visualization_lan}
	\end{subfigure}%

	\begin{subfigure}{.45\textwidth}
		\centering
		\includegraphics[width=\linewidth]{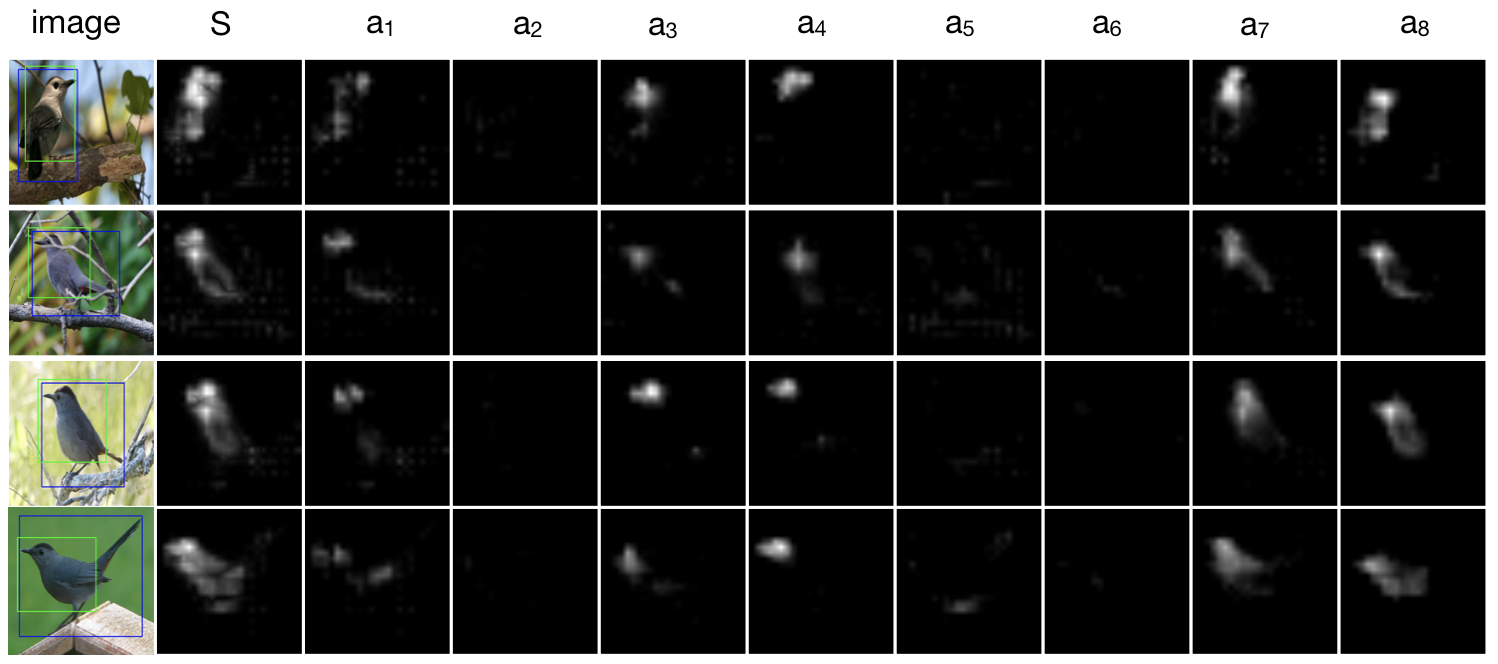}
		\caption{Attention maps generated by WS-BAN. More object's parts are activated and each attention map represents one of the object's parts, such as the head $a_4$, upper part $a_7$ and lower part $a_8$ of the body.}
		\label{fig:visualization_ws_lan}
	\end{subfigure}
	\caption{Visualization of attention maps of bird \textit{Gray Catbird} to demonstrate the effectiveness of weakly supervised attention learning. Blue and green bounding box indicates ground truth and predicted location of object respectively.}
	\label{fig:visualization}
\end{figure}

In Fig~\ref{fig:visualization}, we compare the attention maps of WS-BAN without and with attention regularization and attention dropout to intuitively observe the effectiveness of weakly supervised attention learning. We set the number of attention maps to 8. $a_k$ indicates $k_{th}$ attention map and $S$ represents the mean attention maps. WS-BAN with attention learning is more accurate in localizing the whole object, and each of their attention maps is activated in the same location of object's parts.

\section{Conclusion}
\label{sec:conclusion}
In this paper, we propose a novel discriminative part localization and local feature extraction method to solve the fine-grained visual classification problem. By bilinear attention pooling, we represent object by discriminative part feature matrix. Followed by weakly supervised attention learning, including attention regularization and attention dropout, we guide each attention map to focus on one of the object's parts and encourage multiple attention. We finally achieve the state-of-the-art performance in fine-grained visual classification datasets. In the future, we will extend the bilinear attention pooling method to other tasks, such as object detection and object segmentation.

\bibliographystyle{named}
\bibliography{ijcai19}

\end{document}